\documentclass[11pt,a4paper]{article}
\usepackage[hyperref]{emnlp-ijcnlp-2019}
\usepackage{times}
\usepackage{latexsym}

\usepackage{url}
\usepackage{multirow}
\usepackage{amsmath}
\usepackage{graphicx}
\usepackage{booktabs}
\usepackage{CJKutf8}
\usepackage{linguex}
\usepackage{cgloss4e}

\usepackage{color}

\aclfinalcopy 

\title{Minimally Supervised Learning of Affective Events\\ Using Discourse Relations}

\author{
    Jun Saito \And 
    Yugo Murawaki \\
    Graduate School of Informatics, Kyoto University \\
    Yoshida-honmachi, Sakyo-ku, Kyoto, 606-8501, Japan \\
    {\tt \{saito, murawaki, kuro\}@nlp.ist.i.kyoto-u.ac.jp} \And
    Sadao Kurohashi
  }

\date{}

\begin{document}
\maketitle
\begin{abstract}
Recognizing affective events that trigger positive or negative sentiment has a wide range of natural language processing applications but remains a challenging problem mainly because the polarity of an event is not necessarily predictable from its constituent words.
In this paper, we propose to propagate affective polarity using discourse relations.
Our method is simple and only requires a very small seed lexicon and a large raw corpus.
Our experiments using Japanese data show that our method learns affective events effectively without manually labeled data.
It also improves supervised learning results when labeled data are small.
\end{abstract}

\begin{CJK}{UTF8}{ipxm}

\section{Introduction}

Affective events \cite{ding} are events that typically affect people in positive or negative ways.
For example, getting money and playing sports are usually positive to the experiencers; catching cold and losing one's wallet are negative.
Understanding affective events is important to various natural language processing (NLP) applications such as dialogue systems \cite{dialog}, question-answering systems \cite{qa}, and humor recognition \cite{humor}.
In this paper, we work on recognizing the polarity of an affective event that is represented by a score ranging from $-1$ (negative) to $1$ (positive).
\par
Learning affective events is challenging because, as the examples above suggest, the polarity of an event is not necessarily predictable from its constituent words.
Combined with the unbounded combinatorial nature of language, the non-compositionality of affective polarity entails the need for large amounts of world knowledge, which can hardly be learned from small annotated data.
\par
\begin{figure*}[ht]
\centering
\includegraphics[width=\textwidth]{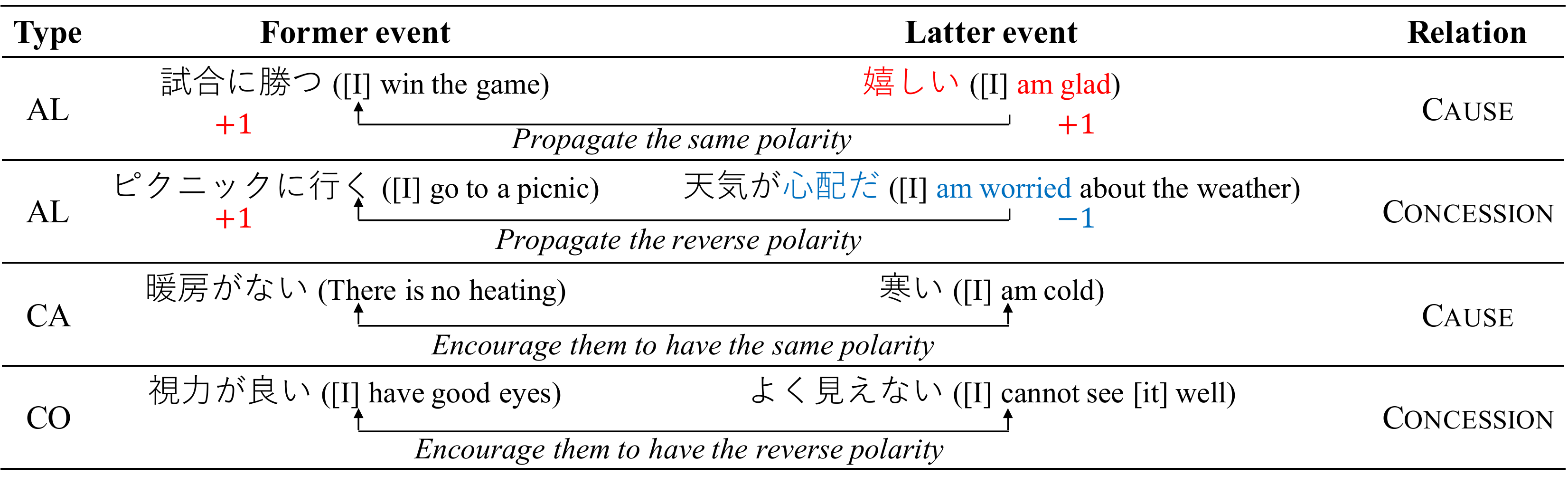}
\caption{
An overview of our method.
We focus on pairs of events, the \textbf{former events} and the \textbf{latter events}, which are connected with a discourse \textbf{relation}, \textsc{Cause} or \textsc{Concession}.
Dropped pronouns are indicated by brackets in English translations.
We divide the event pairs into three \textbf{types}: \textbf{AL}, \textbf{CA}, and \textbf{CO}.
In AL, the polarity of a latter event is automatically identified as either positive or negative, according to the seed lexicon (the positive word is colored red and the negative word blue).
We propagate the latter event's polarity to the former event.
The same polarity as the latter event is used for the discourse relation \textsc{Cause}, and the reversed polarity for \textsc{Concession}.
In CA and CO, the latter event's polarity is not known. 
Depending on the discourse relation, we encourage the two events' polarities to be the same (CA) or reversed (CO).
Details are given in Section \ref{ssec:event_pair}.
}
\label{fig:event_pair}
\vspace{-1.0mm}%
\end{figure*}

In this paper, we propose a simple and effective method for learning affective events that only requires a very small seed lexicon and a large raw corpus.
As illustrated in Figure~\ref{fig:event_pair}, our key idea is that we can exploit discourse relations \cite{pdtb} to efficiently propagate polarity from seed predicates that directly report one's emotions (e.g., ``to be glad'' is positive).
Suppose that events $x_1$ are $x_2$ are in the discourse relation of \textsc{Cause} (i.e., $x_1$ causes $x_2$).
If the seed lexicon suggests $x_2$ is positive, $x_1$ is also likely to be positive because it triggers the positive emotion.
The fact that $x_2$ is known to be negative indicates the negative polarity of $x_1$.
Similarly, if $x_1$ and $x_2$ are in the discourse relation of \textsc{Concession} (i.e., $x_2$ in spite of $x_1$),
the reverse of $x_2$'s polarity can be propagated to $x_1$.
Even if $x_2$'s polarity is not known in advance, we can exploit the tendency of $x_1$ and $x_2$ to be of the same polarity (for \textsc{Cause}) or of the reverse polarity (for \textsc{Concession}) although the heuristic is not exempt from counterexamples.
We transform this idea into objective functions and train neural network models that predict the polarity of a given event.
\par
We trained the models using a Japanese web corpus.
Given the minimum amount of supervision, they performed well.
In addition, the combination of annotated and unannotated data yielded a gain over a purely supervised baseline when labeled data were small.
\par

\section{Related Work}

Learning affective events is closely related to sentiment analysis.
Whereas sentiment analysis usually focuses on the polarity of what are described (e.g., movies), we work on how people are typically affected by events.
In sentiment analysis, much attention has been paid to compositionality.
Word-level polarity~\cite{takamura_spin, wilson, sentiword} and the roles of negation and intensification~\cite{negation, wilson, zhu-negation} are among the most important topics.
In contrast, we are more interested in recognizing the sentiment polarity of an event that pertains to commonsense knowledge (e.g., getting money and catching cold).
\par
Label propagation from seed instances is a common approach to inducing sentiment polarities.
While \citet{takamura_spin} and \citet{Turney} worked on word- and phrase-level polarities, \citet{ding} dealt with event-level polarities.
\citet{takamura_spin} and \citet{Turney} linked instances using co-occurrence information and/or phrase-level coordinations (e.g., ``$A$ and $B$'' and ``$A$ but $B$'').
We shift our scope to event pairs that are more complex than phrase pairs, and consequently exploit discourse connectives as event-level counterparts of phrase-level conjunctions.
\par
\citet{ding} constructed a network of events using word embedding-derived similarities.
Compared with this method, our discourse relation-based linking of events is much simpler and more intuitive.
\par
Some previous studies made use of document structure to understand the sentiment.
\citet{shimizu} proposed a sentiment-specific pre-training strategy using unlabeled dialog data (tweet-reply pairs).
\citet{acp_corpus} proposed a method of building a polarity-tagged corpus (ACP Corpus).
They automatically gathered sentences that had positive or negative opinions utilizing HTML layout structures in addition to linguistic patterns.
Our method depends only on raw texts and thus has wider applicability.

\vspace{-1.5mm}
\section{Proposed Method}
\vspace{-2.5mm}
\subsection{Polarity Function}
\vspace{-1.0mm}

Our goal is to learn the polarity function $p(x)$, which predicts the sentiment polarity score of an event $x$.
We approximate $p(x)$ by a neural network with the following form:
\begin{equation}
\label{pfunc}
\vspace{-1.0mm}%
p(x)=\tanh({\rm Linear}({\rm Encoder}(x))) .
\vspace{-1.0mm}%
\end{equation}
${\rm Encoder}$ outputs a vector representation of the event $x$.
${\rm Linear}$ is a fully-connected layer and transforms the representation into a scalar.
${\rm tanh}$ is the hyperbolic tangent and transforms the scalar into a score ranging from $-1$ to $1$.
In Section~\ref{sec:exp-models}, we consider two specific implementations of ${\rm Encoder}$.

\vspace{-1.5mm}
\subsection{Discourse Relation-Based Event Pairs}
\label{ssec:event_pair}
\vspace{-1.0mm}
Our method requires a very small seed lexicon and a large raw corpus.
We assume that we can automatically extract discourse-tagged event pairs, $(x_{i1}, x_{i2})$ ($i=1, \cdots$) from the raw corpus.
We refer to $x_{i1}$ and $x_{i2}$ as {\it former} and {\it latter} events, respectively.
As shown in Figure~\ref{fig:event_pair}, we limit our scope to two discourse relations: \textsc{Cause} and \textsc{Concession}.

The seed lexicon consists of positive and negative predicates.
If the predicate of an extracted event is in the seed lexicon and does not involve complex phenomena like negation, we assign the corresponding polarity score ($+1$ for positive events and $-1$ for negative events) to the event.
We expect the model to automatically learn complex phenomena through label propagation.
Based on the availability of scores and the types of discourse relations, we classify the extracted event pairs into the following three types.
\vspace{-2.0mm}
\paragraph{AL (Automatically Labeled Pairs)}
The seed lexicon matches (1) the latter event but (2) not the former event,
and (3) their discourse relation type is \textsc{Cause} or \textsc{Concession}.
If the discourse relation type is \textsc{Cause}, the former event is given the same score as the latter.
Likewise, if the discourse relation type is \textsc{Concession}, the former event is given the opposite of the latter's score.
They are used as reference scores during training.
\vspace{-2.0mm}
\paragraph{CA (\textsc{Cause} Pairs)}
The seed lexicon matches neither the former nor the latter event, and their discourse relation type is \textsc{Cause}.
We assume the two events have the same polarities.
\vspace{-2.0mm}
\paragraph{CO (\textsc{Concession} Pairs)}
The seed lexicon matches neither the former nor the latter event, and their discourse relation type is \textsc{Concession}.
We assume the two events have the reversed polarities.

\vspace{-1.5mm}
\subsection{Loss Functions}
\vspace{-1.0mm}
Using AL, CA, and CO data, we optimize the parameters of the polarity function $p(x)$.
We define a loss function for each of the three types of event pairs and sum up the multiple loss functions.

We use mean squared error to construct loss functions.
For the AL data, the loss function is defined as:
\begin{align}
\vspace{-1.0mm}%
\mathcal{L}_{\rm AL} &= \frac{1}{N_{\rm AL}} \sum_{i=1}^{N_{\rm AL}} (r_{i2}-p(x_{i2}))^2 \nonumber \\
&\quad + \lambda_{\rm AL} \frac{1}{N_{\rm AL}}
\sum_{i=1}^{N_{\rm AL}} (r_{i1}-p(x_{i1}))^2,
\vspace{-1.0mm}%
\end{align}
where $x_{i1}$ and $x_{i2}$ are the $i$-th pair of the AL data.
$r_{i1}$ and $r_{i2}$ are the automatically-assigned scores of $x_{i1}$ and $x_{i2}$, respectively.
$N_{\rm AL}$ is the total number of AL pairs, and $\lambda_{\rm AL}$ is a hyperparameter.
\par
For the CA data, the loss function is defined as:
\begin{align}
\vspace{-1.0mm}%
\mathcal{L}_{\rm CA} &= \lambda_{\rm CA} \frac{1}{N_{\rm CA}} \sum_{i=1}^{N_{\rm CA}} (p(y_{i1})-p(y_{i2}))^2 \nonumber \\
&\quad + \mu \frac{1}{N_{\rm CA}} \sum_{i=1}^{N_{\rm CA}} \sum_{u \in \{y_{i1}, y_{i2}\}}^{} (1-p(u)^2).
\vspace{-1.0mm}%
\end{align}
$y_{i1}$ and $y_{i2}$ are the $i$-th pair of the CA pairs.
$N_{\rm CA}$ is the total number of CA pairs.
$\lambda_{\rm CA}$ and $\mu$ are hyperparameters.
The first term makes the scores of the two events closer while the second term prevents the scores from shrinking to zero.
\par
The loss function for the CO data is defined analogously:
\begin{align}
\mathcal{L}_{\rm CO} &= \lambda_{\rm CO} \frac{1}{N_{\rm CO}} \sum_{i=1}^{N_{\rm CO}} (p(z_{i1})+p(z_{i2}))^2 \nonumber \\
&\quad + \mu \frac{1}{N_{\rm CO}} \sum_{i=1}^{N_{\rm CO}} \sum_{u \in \{z_{i1}, z_{i2}\}}^{} (1-p(u)^2).
\end{align}
The difference is that the first term makes the scores of the two events distant from each other.

\vspace{-1.5mm}
\section{Experiments}
\vspace{-2.5mm}
\subsection{Dataset}
\vspace{-1.0mm}
\subsubsection{AL, CA, and CO}
As a raw corpus, we used a Japanese web corpus that was compiled through the procedures proposed by \citet{caseframe}.
To extract event pairs tagged with discourse relations, we used the Japanese dependency parser KNP\footnote{\url{http://nlp.ist.i.kyoto-u.ac.jp/EN/index.php?KNP}} and in-house postprocessing scripts~\cite{event}.
KNP used hand-written rules to segment each sentence into what we conventionally called {\em clauses} (mostly consecutive text chunks), each of which contained one main predicate.
KNP also identified the discourse relations of event pairs if explicit discourse connectives~\cite{pdtb} such as ``ので'' (\textit{because}) and ``のに'' (\textit{in spite of}) were present.
We treated Cause/Reason (原因・理由) and Condition (条件) in the original tagset~\citep{kawahara} as \textsc{Cause} and Concession (逆接)\footnote{
To be precise, this discourse type is semantically broader than Concession and extends to the area of Contrast.
} as \textsc{Concession}, respectively.
Here is an example of event pair extraction.
\vspace{-1.0mm}
\ex. 重大な失敗を犯したので、仕事をクビになった。 \\
Because [I] made a serious mistake, [I] got fired. \par
\vspace{-1.0mm}
From this sentence, we extracted the event pair of ``重大な失敗を犯す'' ([I] make a serious mistake) and ``仕事をクビになる'' ([I] get fired), and tagged it with \textsc{Cause}.
\par
We constructed our seed lexicon consisting of 15 positive words and 15 negative words, as shown in Section~\ref{seed}.
From the corpus of about 100 million sentences, we obtained 1.4 millions event pairs for AL, 41 millions for CA, and 6 millions for CO.
We randomly selected subsets of AL event pairs such that positive and negative latter events were equal in size.
We also sampled event pairs for each of CA and CO such that it was five times larger than AL.
The results are shown in Table~\ref{clause pair}.

\begin{table}[t!]
\begin{center}
\begin{tabular}{lr}
\hline \multicolumn{1}{c}{\bf Type of pairs} & {\bf \# of pairs} \\ \hline
AL (Automatically Labeled Pairs) & 1,000,000 \\
CA (\textsc{Cause} Pairs) & 5,000,000 \\
CO (\textsc{Concession} Pairs) & 5,000,000 \\
\hline
\end{tabular}
\end{center}
\vspace{-3.5mm}%
\caption{
Statistics of the AL, CA, and CO datasets.
\vspace{-1.0mm}%
}
\label{clause pair}
\end{table}
\subsubsection{ACP (ACP Corpus)}
We used the latest version\footnote{The dataset was obtained from Nobuhiro Kaji via personal communication.} of the ACP Corpus~\cite{acp_corpus} for evaluation.
It was used for (semi-)supervised training as well.
Extracted from Japanese websites using HTML layouts and linguistic patterns, the dataset covered various genres.
For example, the following two sentences were labeled positive and negative, respectively:
\vspace{-2.0mm}
\ex. 作業が楽だ。 \\
The work is easy. \par
\ex. 駐車場がない。 \\ 
There is no parking lot. \par
\vspace{-2.0mm}
Although the ACP corpus was originally constructed in the context of sentiment analysis, we found that it could roughly be regarded as a collection of affective events.
We parsed each sentence and extracted the last clause in it.
The train/dev/test split of the data is shown in Table~\ref{acp}.
\begin{table}[t!]
\begin{center}
\begin{tabular}{ccr}
\hline \bf Dataset & \bf Event polarity & \bf \# of events \\ \hline
\multirow{2}{*}{Train} & Positive & 299,834 \\
& Negative & 300,164 \\ \hline
\multirow{2}{*}{Dev} & Positive & 50,118 \\
& Negative & 49,882 \\ \hline
\multirow{2}{*}{Test} & Positive & 50,046 \\
& Negative & 49,954 \\
\hline
\end{tabular}
\end{center}
\vspace{-3.5mm}%
\caption{\label{font-table} Details of the ACP dataset.
\vspace{-1.0mm}%
}
\label{acp}
\end{table}
The objective function for supervised training is: 
\vspace{-1.0mm}%
\begin{align}
\mathcal{L}_{\rm ACP} = \frac{1}{N_{\rm ACP}} \sum_{i=1}^{N_{\rm ACP}} (R_i-p(v_i))^2,
\vspace{-1.5mm}%
\end{align}
where $v_i$ is the $i$-th event,
$R_i$ is the reference score of $v_i$, and
$N_{\rm ACP}$ is the number of the events of the ACP Corpus.
\par
To optimize the hyperparameters, we used the dev set of the ACP Corpus.
For the evaluation, we used the test set of the ACP Corpus.
The model output was classified as positive if $p(x) > 0$ and negative if $p(x) \leq 0$.

\vspace{-1.5mm}
\subsection{Model Configurations} \label{sec:exp-models}
\vspace{-1.0mm}
As for ${\rm Encoder}$, we compared two types of neural networks: BiGRU and BERT.
GRU \cite{gru} is a recurrent neural network sequence encoder.
BiGRU reads an input sequence forward and backward and the output is the concatenation of the final forward and backward hidden states.
\par
BERT \cite{bert} is a pre-trained multi-layer bidirectional Transformer \cite{transformer} encoder.
Its output is the final hidden state corresponding to the special classification tag (\texttt{[CLS]}).
For the details of ${\rm Encoder}$, see Sections~\ref{encoder}.
\par
We trained the model with the following four combinations of the datasets: AL, AL+CA+CO (two proposed models), ACP (supervised), and ACP+AL+CA+CO (semi-supervised).
The corresponding objective functions were:
$\mathcal{L}_{\rm AL}$, $\mathcal{L}_{\rm AL} + \mathcal{L}_{\rm CA} + \mathcal{L}_{\rm CO}$, $\mathcal{L}_{\rm ACP}$, and $\mathcal{L}_{\rm ACP} + \mathcal{L}_{\rm AL} + \mathcal{L}_{\rm CA} + \mathcal{L}_{\rm CO}$.

\vspace{-1.5mm}
\subsection{Results and Discussion}
\vspace{-1.0mm}
\begin{table}[t!]
\begin{center}
\begin{tabular}{llrr}
\hline \multicolumn{1}{c}{\bf Training dataset} & \bf Encoder & \multicolumn{1}{c}{\bf Acc} \\ \hline 
\multirow{2}{*}{AL} & BiGRU & 0.843 \\
& BERT & 0.863 \\
\multirow{2}{*}{AL+CA+CO} & BiGRU & \bf 0.866 \\
& BERT & 0.835 \\ \hline
\multirow{2}{*}{ACP} & BiGRU & 0.919 \\
& BERT & \bf 0.933 \\
\multirow{2}{*}{ACP+AL+CA+CO} & BiGRU & 0.917 \\
& BERT & 0.913 \\ \hline \hline
\multicolumn{2}{l}{Random} & 0.500 \\ 
\multicolumn{2}{l}{Random+Seed} & 0.503 \\ \hline
\end{tabular}
\end{center}
\vspace{-3.5mm}%
\caption{
Performance of various models on the ACP test set.
}
\label{results}
\vspace{-1.0mm}%
\end{table}

\begin{table}[t!]
\begin{center}
\begin{tabular}{llrr}
\hline \multicolumn{1}{c}{\bf Training dataset} & \multicolumn{1}{c}{\bf Encoder} & \multicolumn{1}{c}{\bf Acc} \\ \hline
ACP (6K) & \multirow{2}{*}{BERT} & 0.876 \\
\quad+AL && \bf 0.886 \\ \hline
ACP (6K) & \multirow{2}{*}{BiGRU} & 0.830 \\
\quad+AL+CA+CO && \bf 0.879 \\ \hline
\end{tabular}
\end{center}
\vspace{-3.5mm}%
\caption{
Results for small labeled training data.
Given the performance with the full dataset, we show BERT trained only with the AL data.
}
\label{additional_results}
\vspace{-2.5mm}%
\end{table}

Table~\ref{results} shows accuracy.
As the Random baseline suggests, positive and negative labels were distributed evenly.
The Random+Seed baseline made use of the seed lexicon and output the corresponding label (or the reverse of it for negation) if the event's predicate is in the seed lexicon.
We can see that the seed lexicon itself had practically no impact on prediction.
\par
The models in the top block performed considerably better than the random baselines.
The performance gaps with their (semi-)supervised counterparts, shown in the middle block, were less than 7\%.
This demonstrates the effectiveness of discourse relation-based label propagation.
\par
Comparing the model variants, we obtained the highest score with the BiGRU encoder trained with the AL+CA+CO dataset.
BERT was competitive but its performance went down if CA and CO were used in addition to AL.
We conjecture that BERT was more sensitive to noises found more frequently in CA and CO.
\par
Contrary to our expectations, supervised models (ACP) outperformed semi-supervised models (ACP+AL+CA+CO).
This suggests that the training set of 0.6 million events is sufficiently large for training the models.
For comparison, we trained the models with a subset (6,000 events) of the ACP dataset. 
As the results shown in Table~\ref{additional_results} demonstrate, our method is effective when labeled data are small.
\par
The result of hyperparameter optimization for the BiGRU encoder was as follows:
\begin{align}
\vspace{-1.0mm}
\lambda_{\rm AL} = 1, \, \lambda_{\rm CA} = 0.35, \, \lambda_{\rm CO} = 1, \,  \mu = 0.5. \nonumber
\vspace{-1.0mm}
\end{align}
As the CA and CO pairs were equal in size (Table \ref{clause pair}), $\lambda_{\rm CA}$ and $\lambda_{\rm CO}$ were comparable values.
$\lambda_{\rm CA}$ was about one-third of $\lambda_{\rm CO}$, and this indicated that the CA pairs were noisier than the CO pairs.
A major type of CA pairs that violates our assumption was in the form of ``$\textit{problem}_{\text{negative}}$ causes $\textit{solution}_{\text{positive}}$'':
\vspace{-1.0mm}
\ex. (悪いところがある, よくなるように努力する) \\
(there is a bad point, [I] try to improve [it]) \par
\vspace{-1.0mm}
The polarities of the two events were reversed in spite of the \textsc{Cause} relation, and this lowered the value of $\lambda_{\rm CA}$.
\par
Some examples of model outputs are shown in Table~\ref{samples}.
The first two examples suggest that our model successfully learned negation without explicit supervision.
Similarly, the next two examples differ only in voice but the model correctly recognized that they had opposite polarities.
The last two examples share the predicate ``落とす" (drop) and only the objects are different.
The second event ``肩を落とす" (lit. drop one's shoulders) is an idiom that expresses a disappointed feeling.
The examples demonstrate that our model correctly learned non-compositional expressions.
\begin{table}[t!]
\begin{center}
\begin{tabular}{lr}
\hline \multicolumn{1}{c}{\bf Input event} & \multicolumn{1}{c}{\bf Polarity} \\ \hline
道に迷う ([I] get lost) & -0.771 \\
道に迷わない ([I] don't get lost) & 0.835 \\ \hline
笑う ([I] laugh) & 0.624 \\
笑われる ([I] am laughed at) & -0.687 \\ \hline
脂肪を落とす ([I] lose body fat) & 0.452  \\
肩を落とす ([I] feel disappointed) & -0.653 \\ \hline
\end{tabular}
\end{center}
\vspace{-3.5mm}%
\caption{
Examples of polarity scores predicted by the BiGRU model trained with AL+CA+CO.
}
\label{samples}
\vspace{-2.5mm}%
\end{table}

\vspace{-1.5mm}
\section{Conclusion}
\vspace{-2.5mm}
In this paper, we proposed to use discourse relations to effectively propagate polarities of affective events from seeds.
Experiments show that, even with a minimal amount of supervision, the proposed method performed well.
\par
Although event pairs linked by discourse analysis are shown to be useful, they nevertheless contain noises.
Adding linguistically-motivated filtering rules would help improve the performance.

\section*{Acknowledgments}
We thank Nobuhiro Kaji for providing the ACP Corpus and Hirokazu Kiyomaru and Yudai Kishimoto for their help in extracting event pairs.
This work was partially supported by Yahoo! Japan Corporation.

\bibliography{sentiment_learning}
\bibliographystyle{acl_natbib}

\clearpage

\appendix
\section{Appendices}
\subsection{Seed Lexicon} \label{seed}
\paragraph{Positive Words}
喜ぶ (rejoice), 嬉しい (be glad), 楽しい (be pleasant), 幸せ (be happy), 感動 (be impressed), 興奮 (be excited), 懐かしい (feel nostalgic), 好き (like), 尊敬 (respect), 安心 (be relieved), 感心 (admire), 落ち着く (be calm), 満足 (be satisfied), 癒される (be healed), and スッキリ (be refreshed).
\paragraph{Negative Words}
怒る (get angry), 悲しい (be sad), 寂しい (be lonely), 怖い (be scared), 不安 (feel anxious), 恥ずかしい (be embarrassed), 嫌 (hate), 落ち込む (feel down), 退屈 (be bored), 絶望 (feel hopeless), 辛い (have a hard time), 困る (have trouble), 憂鬱 (be depressed), 心配 (be worried), and 情けない (be sorry).

\subsection{Settings of Encoder} \label{encoder}
\paragraph{BiGRU}
The dimension of the embedding layer was 256. 
The embedding layer was initialized with the word embeddings pretrained using the Web corpus.
The input sentences were segmented into words by the morphological analyzer Juman++.\footnote{\url{http://nlp.ist.i.kyoto-u.ac.jp/EN/index.php?JUMAN++}}
The vocabulary size was 100,000.
The number of hidden layers was 2.
The dimension of hidden units was 256.
The optimizer was Momentum SGD \cite{msgd}.
The mini-batch size was 1024.
We ran 100 epochs and selected the snapshot that achieved the highest score for the dev set.
\paragraph{BERT}
We used a Japanese BERT model\footnote{\url{http://nlp.ist.i.kyoto-u.ac.jp/index.php?BERT\%E6\%97\%A5\%E6\%9C\%AC\%E8\%AA\%9EPretrained\%E3\%83\%A2\%E3\%83\%87\%E3\%83\%AB} (in Japanese)} pretrained with Japanese Wikipedia.
The input sentences were segmented into words by Juman++, and words were broken into subwords by applying BPE \cite{bpe}.
The vocabulary size was 32,000.
The maximum length of an input sequence was 128.
The number of hidden layers was 12.
The dimension of hidden units was 768.
The number of self-attention heads was 12.
The optimizer was Adam \cite{adam}.
The mini-batch size was 32.
We ran 1 epoch.

\end{CJK}
\end{document}